\documentclass[Afour,sageh,times]{sagej}

\usepackage{moreverb,url}
\usepackage{pgfplots}
\usepackage{subcaption}
\usepackage{dirtree}
\usepackage{setspace}

\usepackage[colorlinks,bookmarksopen,bookmarksnumbered,citecolor=red,urlcolor=red]{hyperref}

\newcommand\BibTeX{{\rmfamily B\kern-.05em \textsc{i\kern-.025em b}\kern-.08em
T\kern-.1667em\lower.7ex\hbox{E}\kern-.125emX}}

\def\volumeyear{2021}

\begin{document}

\runninghead{Kramer, Harlow, Williams, and Heckman}

\title{ColoRadar: The Direct 3D Millimeter Wave Radar Dataset}

\author{Andrew Kramer\affilnum{1}, Kyle Harlow\affilnum{1}, Christopher Williams\affilnum{2}, and Christoffer Heckman\affilnum{1}} 

\affiliation{
    \affilnum{1}
    Department of Computer Science,  University of Colorado Boulder, 1111 Engineering Dr, Boulder Colorado USA \\
    \affilnum{2}
    Department of Aerospace Engineering, University of Colorado Boulder, 3775 Discovery Dr,
    Boulder Colorado USA    
}

\corrauth{Andrew Kramer, University of Colorado Boulder, 1111 Engineering Dr, Boulder Colorado USA}

\email{andrew.kramer@colorado.edu}

\begin{abstract}
Millimeter wave radar is becoming increasingly popular as a sensing modality for robotic mapping and state estimation. However, there are very few publicly available datasets that include dense, high-resolution millimeter wave radar scans and there are none focused on 3D odometry and mapping. In this paper we present a solution to that problem. The ColoRadar dataset includes 3 different forms of dense, high-resolution radar data from 2 FMCW radar sensors as well as 3D lidar, IMU, and highly accurate groundtruth for the sensor rig's pose over approximately 2 hours of data collection in highly diverse 3D environments. 
\end{abstract}


\maketitle

\section{Introduction}
\subsection{Motivation and Contribution}

Robots are being deployed in ever more challenging roles. In many environments GPS is unavailable so autonomous robots must be able to localize and map using only their onboard sensors. These tasks are normally accomplished using some combination of visual, lidar, and inertial sensors  \cite{cartographer,okvis_2013,okvis_2015}. However, the quality of the estimates from visual and lidar methods is severely handicapped in so-called visually degraded environments (VDEs). These include darkness, fog, smoke, and similar conditions that adversely affect sensing in the visible or near-visible spectrum. Such conditions are common in a variety of scenarios that are of keen interest for robotics research including disaster response and subterranean exploration. 

Millimeter wave radars, of the type used in advanced driver assistance systems (ADAS), are an attractive option for perception in VDEs. These sensors do not require ambient light to operate and are unaffected by airborne particulates like smoke and fog. Additionally, automotive-grade system-on-chip (SoC) radars are lightweight and have low power requirements, making them attractive for vehicles with limited payload and power capacity, e.g.\ micro aerial vehicles (MAVs). Because of these attributes the number of robotic perception papers using automotive millimeter wave radar sensors has increased drastically in the last few years \cite{schuster_robust_2016,schuster_landmark_2016,Rapp2017,kramer_icra_2020,Lu2020,XiaoxuanLu}. 

However, several factors make the development of new radar-based methods extremely difficult. First, there are very few radar sensors available that are suitable for robotics work. Large 2D scanning sensors \cite{cen_precise_2018,cen_radar-only_2019,Lakshay2020,Saftescu2020}, and small 3D automovotive sensors dominate the market \cite{schuster_robust_2016,schuster_landmark_2016,Rapp2017,kramer_icra_2020,Lu2020,XiaoxuanLu}.  Second, almost all of the available radar sensors implement their own signal processing and target detection systems onboard \cite{TI_mmWave_fundamentals}. These are well adapted for ADAS but ill-suited to robotic perception tasks \cite{Lu2020}. It would be ideal if researchers could implement their own target detection and/or signal processing algorithms, but most sensors do not allow the user access to their full data stream and those that do are not designed for streaming in real time. Hacking these sensors to stream data in real time takes both domain-specific expertise and a large time investment which is prohibitive for most researchers. The final nail in the coffin for most nascent radar-based robotics projects is there are no publicly available robotics datasets that include this data.

To remove these obstacles we present the first multi-sensor dataset designed for 6DoF radar-based state estimation and mapping. The dataset includes dense 4D data from millimeter wave radar sensors, 3D lidar pointclouds, IMU measurements, and groundtruth pose information. The dataset is available for download at \href{https://arpg.colorado.edu/coloradar}{\color{blue}\underline{arpg.colorado.edu/coloradar}}. We include radar data in 3 different forms, providing entry points of varying sophistication and difficulty for users. First, we provide unprocessed ADC samples from each radar measurement. Second, we provide dense 3D heatmaps for each radar measurement generated from signal processing on the raw ADC samples. Lastly, we provide point targets derived from the sensor's onboard signal processing and target detection process. Although as previously stated, these point targets are not ideal for robotic state estimation and perception applications, we include point targets because several recent methods have made use of them \cite{kramer_icra_2020,Lu2020,XiaoxuanLu}. So including point targets provides a familiar entry point for many researchers. 


\subsection{Related Work}
\label{sec:related-work}
While there exist at least three publicly available robotics datasets containing radar data \cite{RadarRobotCarDatasetICRA2020,caesar2020nuscenes,Leung2017ChileanMine}, each dataset is limited in some manner. The Oxford Radar RobotCar dataset \cite{RadarRobotCarDatasetICRA2020} and NuScenes dataset \cite{caesar2020nuscenes} both focus on autonomous road vehicles, limiting sensor movement largely to a 2D horizontal plane. In contrast, the Radar RobotCar dataset uses a different style of radar sensor, the NavTech CTS350-X, which provides dense 2D scans in the horizontal plane only. The NavTech sensor's 2D output makes the Radar RobotCar dataset unsuitable for developing methods for 3D environments. Also, the NavTech sensor weighs 6kg and draws 24W, making it impractical for small, agile autonomous vehicles like MAVs. The Chilean underground mine dataset provides a different environment, it also uses a 2D scanning radar and is mostly confined to 2D movement. \cite{Leung2017ChileanMine}. Lastly, a major limitation common to all of these datasets is that they do not include the raw ADC sample data from their radar sensors. This means users of these datasets must be satisfied with the signal processing and target detection algorithms implemented onboard their sensors. These algorithms are often very ill-suited to robotics tasks as previously stated.


The Euroc MAV dataset \cite{euroc} and the UZH FPV dataset \cite{Delmerico19icra} are examples of datasets that are useful for the development of algorithms for small, agile robots operating in 3D environments. However, these are visual-inertial datasets that do not include radar. Currently there are no similar datasets that include radar data in any form.

Like the Euroc MAV and UZH FPV datasets \cite{euroc, Delmerico19icra}, our dataset includes both slow and rapid 6DoF motion. This means our dataset can be used in the development of state estimation and mapping methods for small, agile robots that operate in diverse 3D environments. Also, our dataset includes both raw ADC data and dense images from multiple types of 4D radar sensors. This allows users to develop new methods for signal processing and target detection which are specifically designed for robotics tasks. Lastly, we include data gathered in highly diverse indoor, outdoor, and subterranean environments. This allows users to develop methods for a wide variety of environments and scenarios. These are clear advantages over the currently-available radar datasets.

\section{Sensor Platform} 
\begin{figure}[h!]
 \centering
 \includegraphics[width=0.85\linewidth]{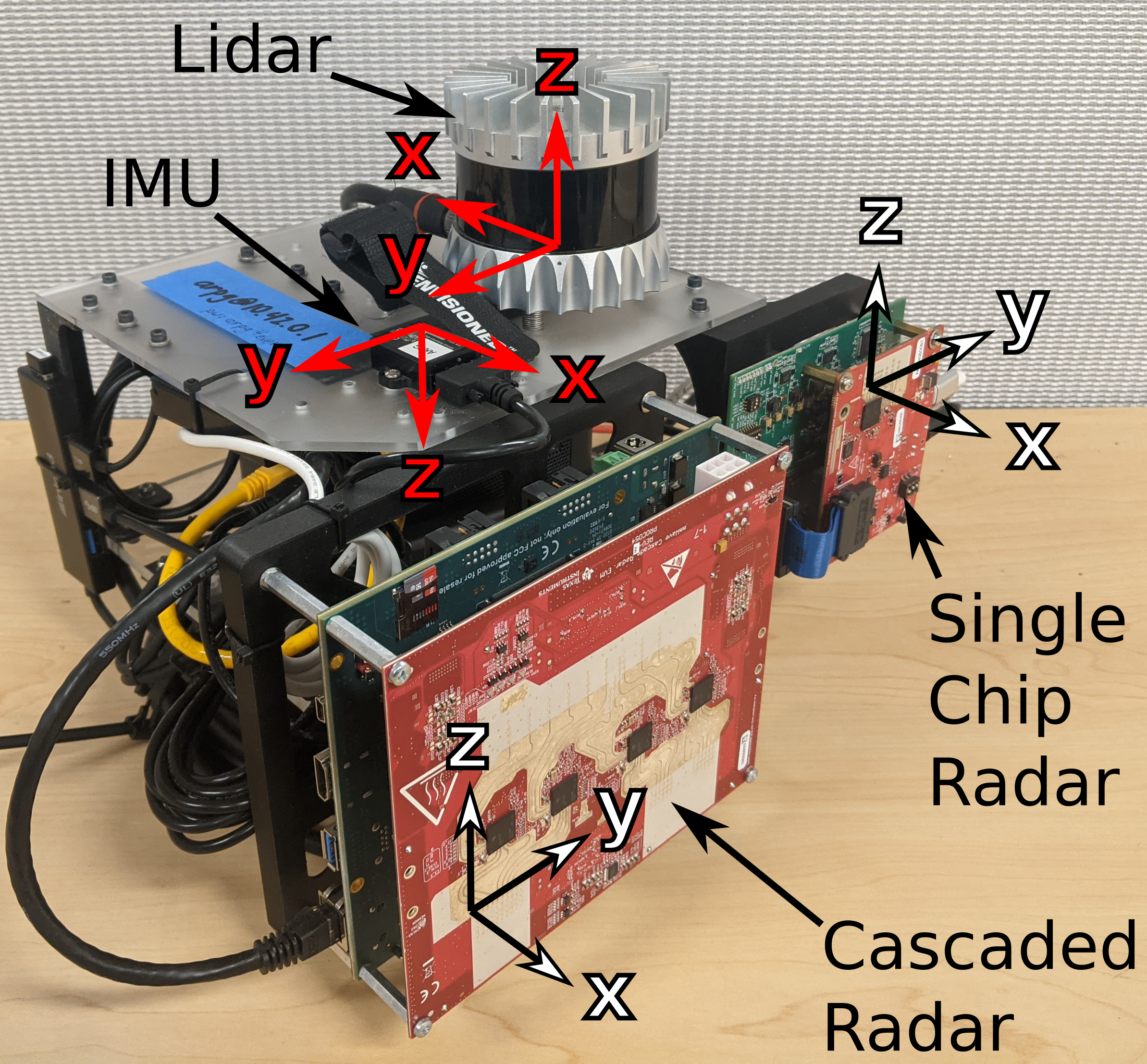}
 \caption{The sensor rig used to collect our dataset. Each sensor is labeled and its local coordinate frame is noted.}
 \label{fig:sensor_rig}
\end{figure}

Our sensor platform is shown in Figure \ref{fig:sensor_rig}. It includes
\begin{itemize}
    \item Cascaded Imaging Radar Sensor: Texas Instruments MMWCAS-RF-EVM  (details in Table \ref{tab:radar_params})
    \item Single Chip Radar Sensor: Texas Instruments AWR1843BOOST-EVM paired with a DCA1000-EVM for raw data capture (details in Table \ref{tab:radar_params})
    \item Lidar Sensor: Ouster OS1; 10Hz; 64 beams; $0.01^\circ$ angular accuracy; angular resolution $0.35^\circ$ horizontal, $0.7^\circ$ vertical; 3cm range accuracy; field of view $360^\circ$ horizontal, $45^\circ$ vertical; max range 120m; 65,536 points per scan
    \item IMU: Lord Microstrain 3DM-GX5-25; 300Hz
\end{itemize}

Note our sensor rig includes two Texas Instruments radar sensors. The single chip sensor is based on the AWR1843 radar on a chip. The cascaded sensor uses four AWR2243 radar chips which function as a single radar. The cascade sensor provides far better angular resolution at the expense of lower framerate, higher data rate, and higher power draw. But because both sensors use a frequency modulated continuous (FMCW) waveform the same signal processing and target detection algorithms may be used for either. The salient parameters of the two sensors are outlined in Table \ref{tab:radar_params}. Note for radar sensors ``resolution'' refers to the minimum separation between targets required to distinguish those targets in the sensor's output. This is distinct from camera resolution, which is simply the dimensions of the camera's output. For sensor control and data logging the sensor rig is equipped with an Intel NUC computer with an i7 processor, 16GB of memory, and 500GB of storage. The computer runs Ubuntu 18.04 and we use ROS Melodic \cite{ros} to control the sensors and log their data streams.

\begin{table}
\centering
 \begin{tabular}{| p{3.0cm} p{2.0cm} p{2.0cm}  |} 
 \hline
  & Single Chip Radar & Cascaded Radar \\ [0.5ex]
\hline\hline 
 Framerate & 10Hz & 5Hz \\ [0.5ex]
 \hline
 Frequency & 77GHz & 77GHz \\ [0.5ex]
 \hline
 Waveform & FMCW & FMCW \\ [0.5ex]
 \hline
 TX antennas & 3 & 12 \\  [0.5ex] 
 \hline
 RX antennas & 4 & 16 \\ [0.5ex]
 \hline
 Range resolution & 0.125m & 0.117m \\ [0.5ex]
 \hline
 Max range & 12m & 15m \\ [0.5ex]
 \hline
 Doppler velocity resolution & 0.04m/s & 0.254m/s \\ [0.5ex]
 \hline
 Max Doppler velocity & 2.56m/s & 2.02m/s \\ [0.5ex]
 \hline
 Azimuth resolution & $11.3^\circ$ & $1.05^\circ$ \\ [0.5ex]
 \hline
 Elevation resolution & $45^\circ$ & $22.5^\circ$ \\ [0.5ex]
 \hline
 Power consumption & 2W & 9.5W \\ [0.5ex]
 \hline
 Data rate & 63Mbps & 126Mbps \\ [0.5ex]
 \hline
\end{tabular}
\caption{Parameters of the radar sensors} \label{tab:radar_params}
\end{table}

\section{Dataset}

\subsection{Radar Signal Processing}

To understand the different types of radar data provided in our dataset, it is necessary to give a brief description of the typical signal processing pipeline for multi-input-multi-output (MIMO) frequency modulated continuous wave (FMCW) radar sensors. 

\subsubsection{Range-Doppler Velocity Processing}

For simplicity we will initially consider the data format for a radar with a single receive (RX) and a single transmit (TX) antenna. Each radar measurement consists of repetitions of the same unit, referred to as a ``chirp.'' For each chirp, the TX antenna emits a pulse of RF energy with linearly increasing frequency. This pulse is reflected by objects in the environment and these reflections are detected by the RX antenna. The signal detected by the RX antenna is down-converted by mixing with the original transmitted signal, then this intermediate frequency (IF) signal is sampled by an analog-to-digital converter (ADC), creating an array of $N_S$ complex samples. The sampling done over a single chirp is referred to as ``fast-time.'' The chirp is repeated $N_D$ times, and the received signals from each chirp are stacked to form a 2D array referred to as a ``frame,'' shown in Figure \ref{fig:3D_data_cube}. The height dimension of the frame is referred to as ``slow time.''


\begin{figure}[h!]
 \centering
 \includegraphics[width=0.9\linewidth]{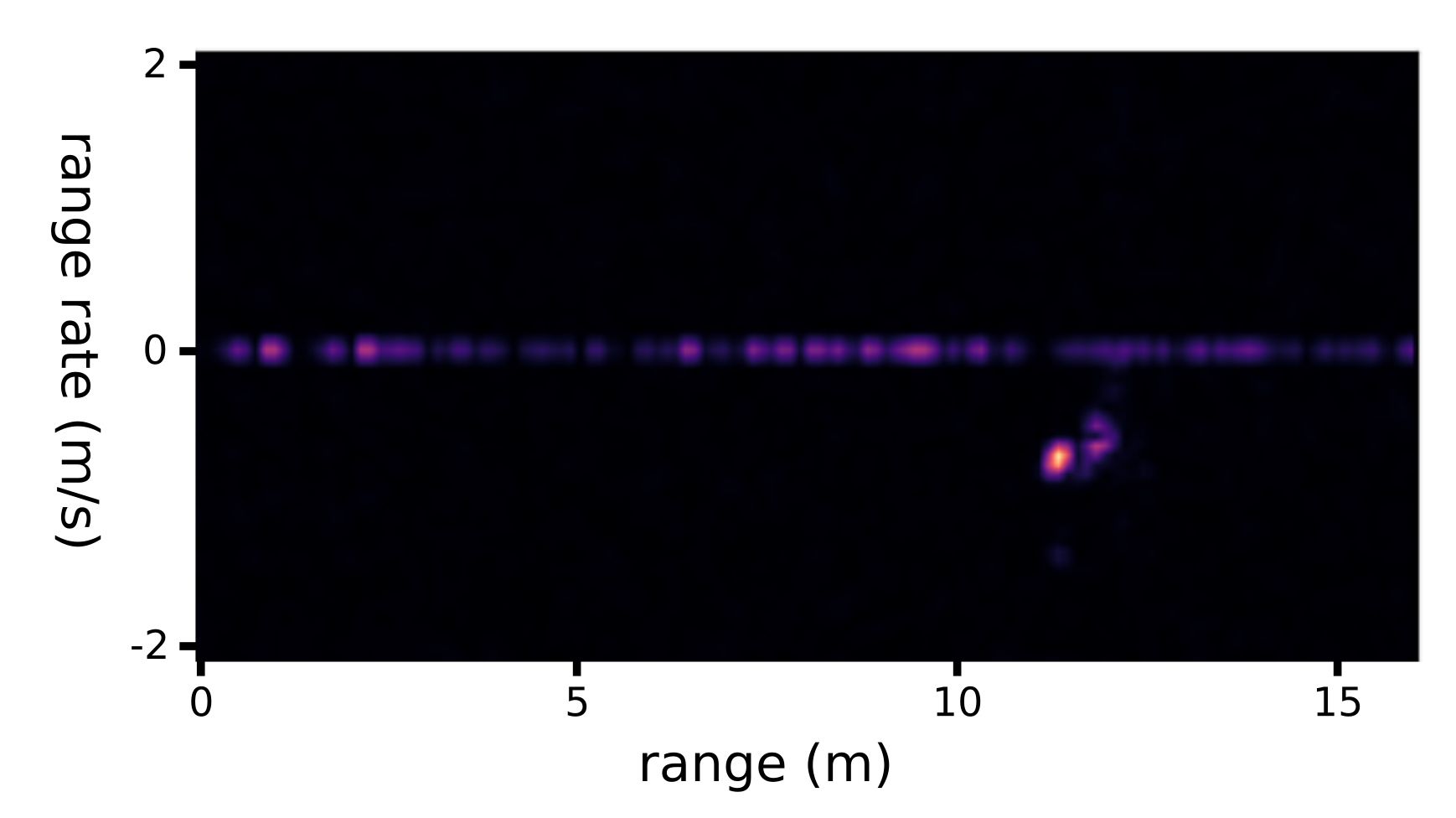}
 \caption{Example result of the 2D FFT run on a frame of data for a single antenna. Note the horizontal band of returns at Doppler velocity bin 0 represent static objects. The intensity peak at range 11 and range rate -0.8 indicates an object at 11m with range decreasing at 0.8m/s.}
 \label{fig:range_doppler_plot}
\end{figure}

Complex fast Fourier transforms (FFTs) are then performed across the fast time and slow time dimensions of the array, resulting in a profile of the received power in fast and slow time. Luckily, there are linear relationships between fast time and range, and between slow time and range rate. So the FFT output can be easily converted to a plot of received power in range and range rate. Figure \ref{fig:range_doppler_plot} shows the result of range-Doppler velocity processing processing for an example frame.

\subsubsection{The MIMO Antenna Array}

MIMO sensors, such as those used in our dataset, use $N_T$ TX and $N_R$ RX antennas. The signals transmitted by each TX antenna are multiplexed in the time domain, which allows us to split the signals detected by the RX antennas into the components caused by each transmitted signal. This means for each frame of radar data we have $N_TN_R$ unique received signals. 

\begin{figure}
\centering
\includegraphics[width=0.85\linewidth]{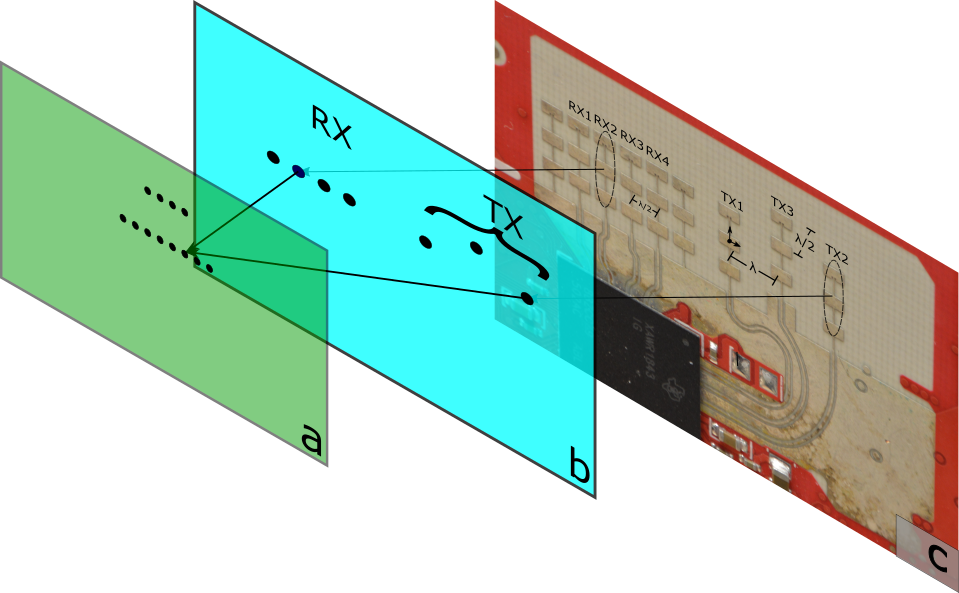}
\caption{Layer (a) shows an example of virtual antenna array synthesis created from the logical antenna layout (b) from a TI AWR1843 patch antenna array (c).} 
\label{fig:antenna_arrays}
\end{figure}

Each RX-TX antenna pair can be considered as a ``virtual'' antenna element, the location of which is the centroid of the physical RX and TX antennas associated with it. 
Figure \ref{fig:antenna_arrays} demonstrates how the spatial layout of antennas in a MIMO array translates to a virtual antenna pattern. Because these virtual antennas are spatially distributed, the relative phase of the signal received by each antenna will depend on the angle of arrival of that signal. This is what allows us to do the angle of arrival processing step.

\subsubsection{Angle Of Arrival Processing}

\begin{figure}[h!]
 \centering
 \includegraphics[width=0.65\linewidth]{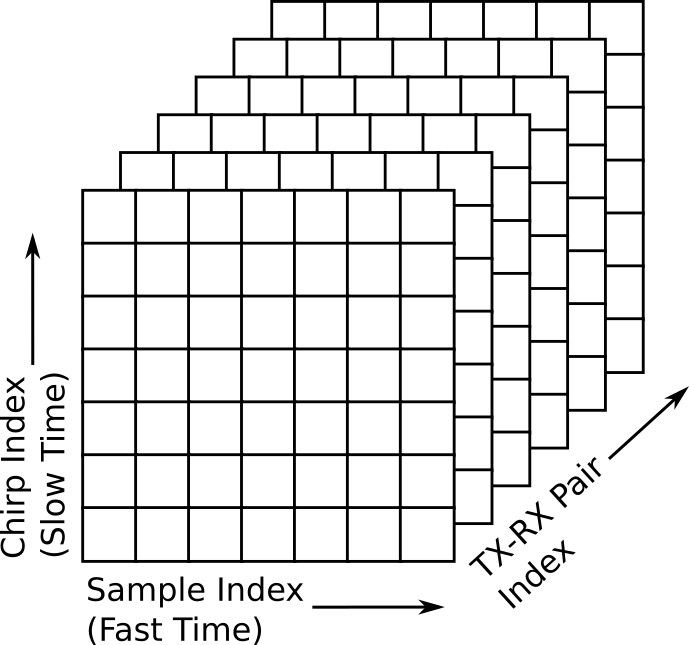}
 \caption{Data cube format for a complete frame of radar data for a multiple virtual antennas. Each box represents a single complex ADC sample and each 2D matrix represents the samples for a full frame of data for a single virtual antenna.}
 \label{fig:3D_data_cube}
\end{figure}

\begin{figure}[h!]
 \centering
 \includegraphics[width=0.45\linewidth]{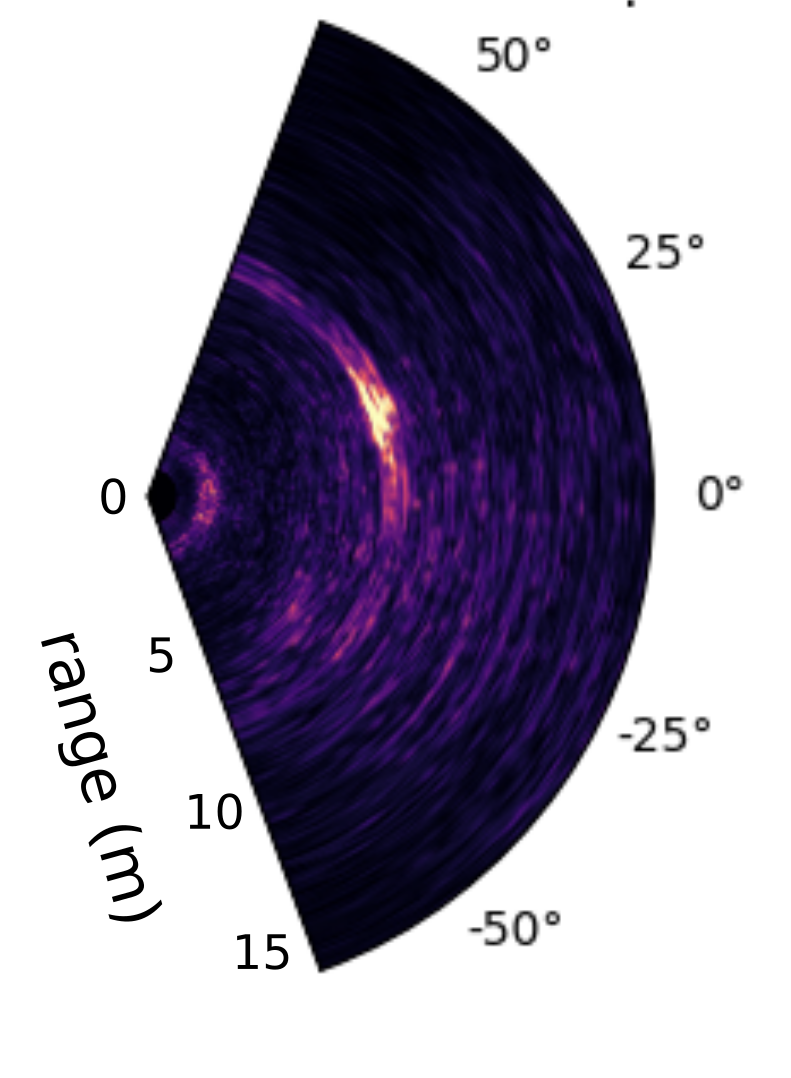}
 \caption{Example result of the angle of arrival FFT. Note the clear intensity peak visible at 8m range and $+25^\circ$ azimuth. The Doppler velocity dimension is not shown for simplicity.}
 \label{fig:aoa_plot}
\end{figure}

For a MIMO antenna array, range-Doppler velocity processing is performed on the samples from each virtual antenna element. The outputs from the range-Doppler velocity processing step are then stacked to form a complex data cube as shown in Figure \ref{fig:3D_data_cube}. Then a third FFT is performed across the virtual antenna dimension, resulting in a heat map of the signal power received by the sensor in the range, Doppler velocity, and azimuth angle dimensions. An example heat map is shown in Figure \ref{fig:aoa_plot}.

By using an antenna array with virtual elements distributed in two dimensions rather than just one as shown in Figure \ref{fig:antenna_arrays}, this process can be extended one step further. Performing fast Fourier transforms across both dimensions of the antenna array results in a 4D heat map in the range, Doppler velocity, azimuth, and elevation dimensions. 

In many off-the-shelf automotive radar sensors an additional target detection step is performed. A constant false alarm rate (CFAR) target detector \cite{cfar} is used to detect peaks that stand out prominently from their surroundings in the range-Doppler velocity-angle heat map. The locations and intensities of these peaks are then returned to the user as a set of point targets. This reduces the size of each measurement substantially, but it also discards a large amount of potentially valuable information. The target detection process also introduces noise because a portion of the detected targets will be false alarms.

\subsection{Data Description}

\begin{figure*}[ht]
\centering
\begin{tabular}{cccc}
\includegraphics[width=0.3\textwidth]{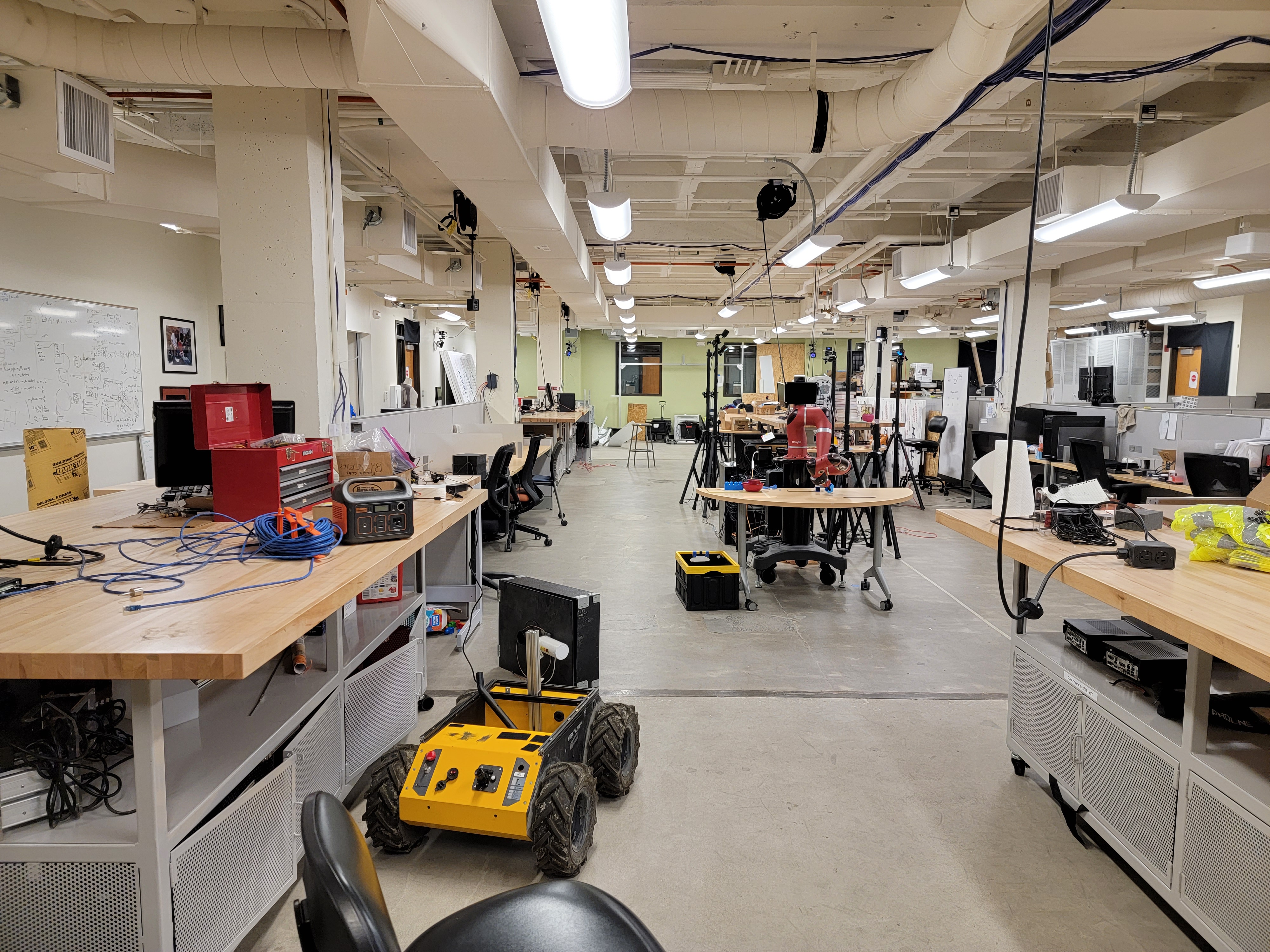} & \includegraphics[width=0.3\textwidth]{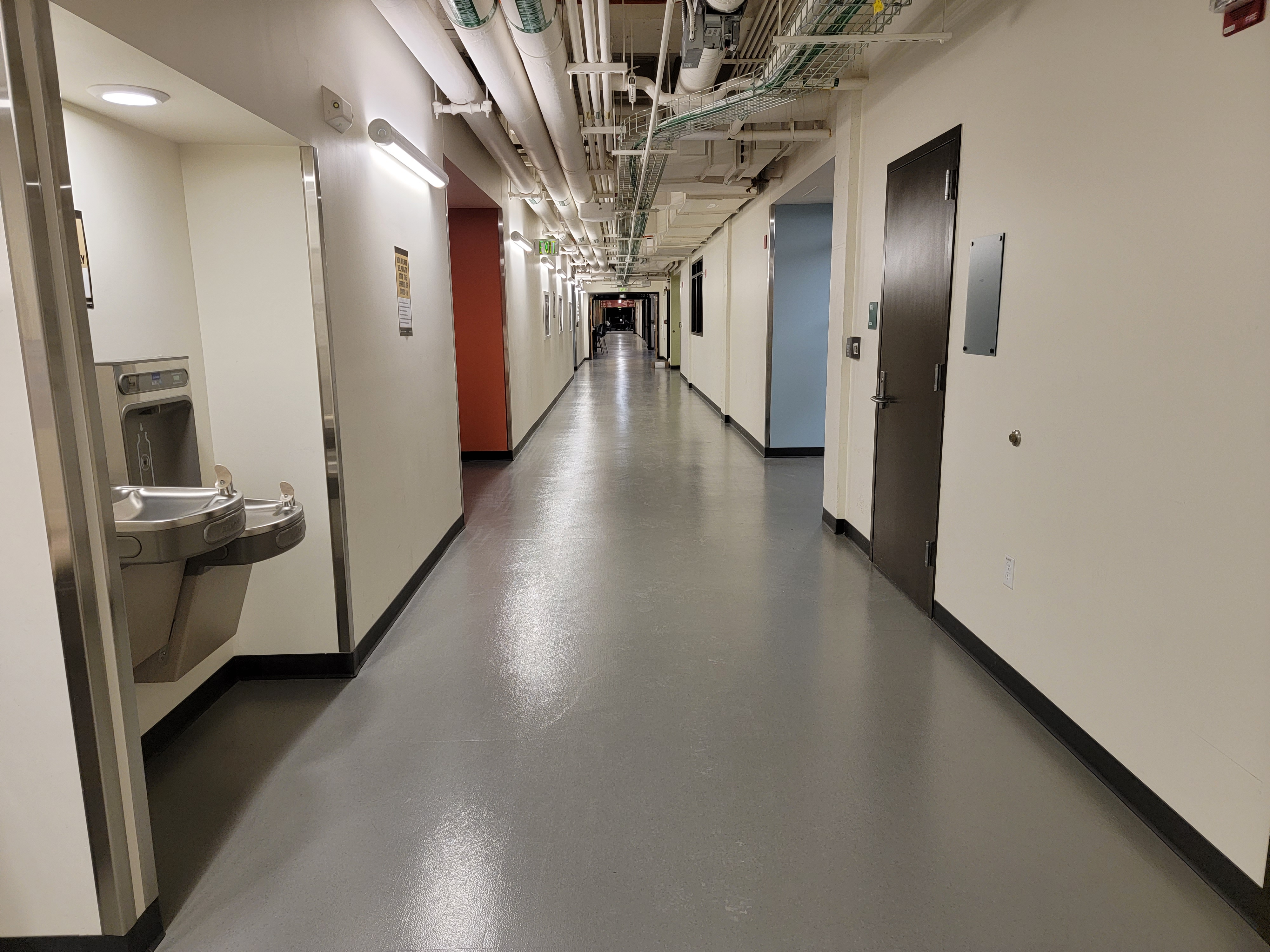} & \includegraphics[width=0.3\textwidth]{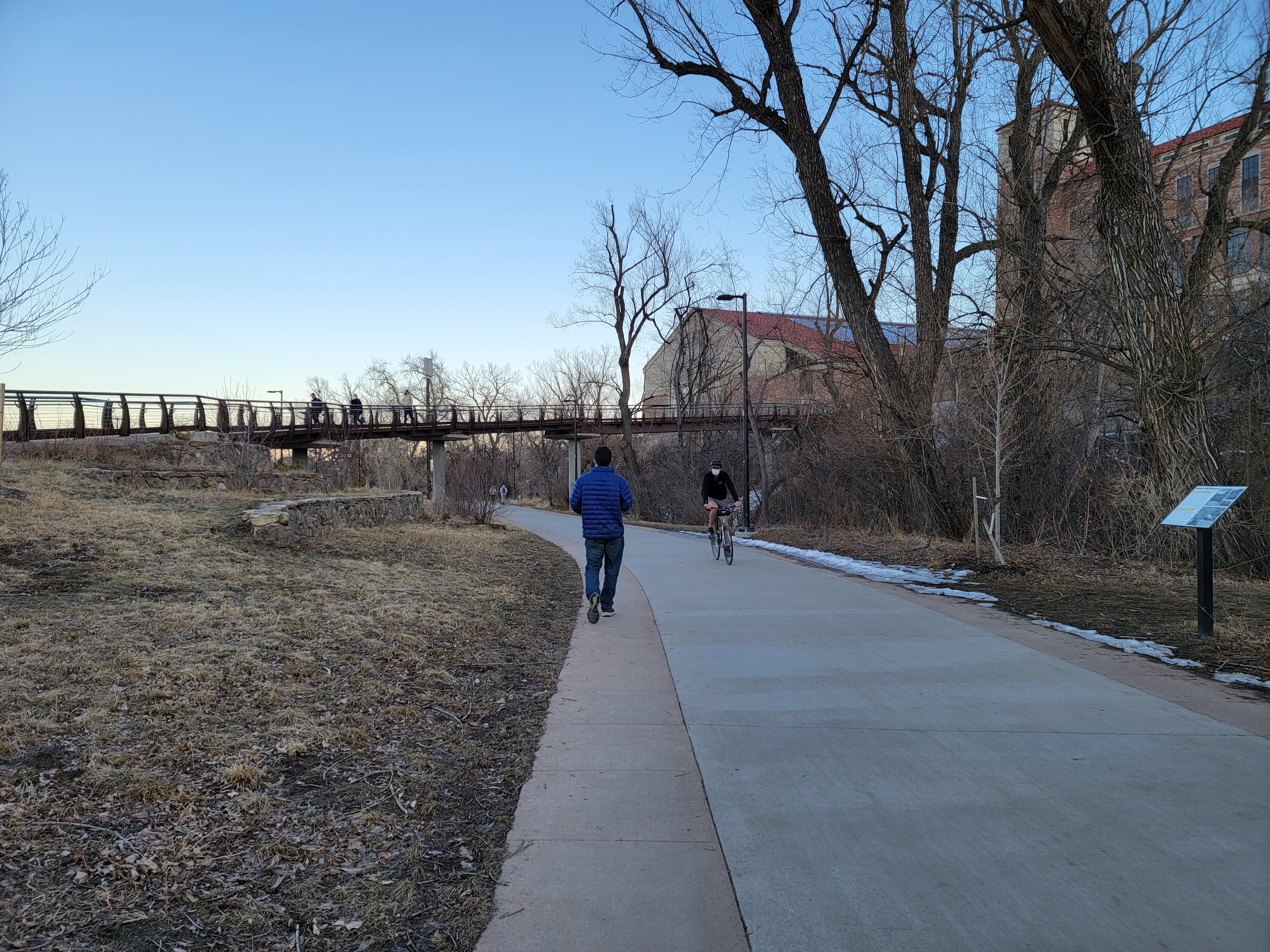}\\
\textbf{(a)} & \textbf{(b)} & \textbf{(c)}\\[6pt]
\end{tabular}
\begin{tabular}{cccc}
\includegraphics[width=0.3\textwidth]{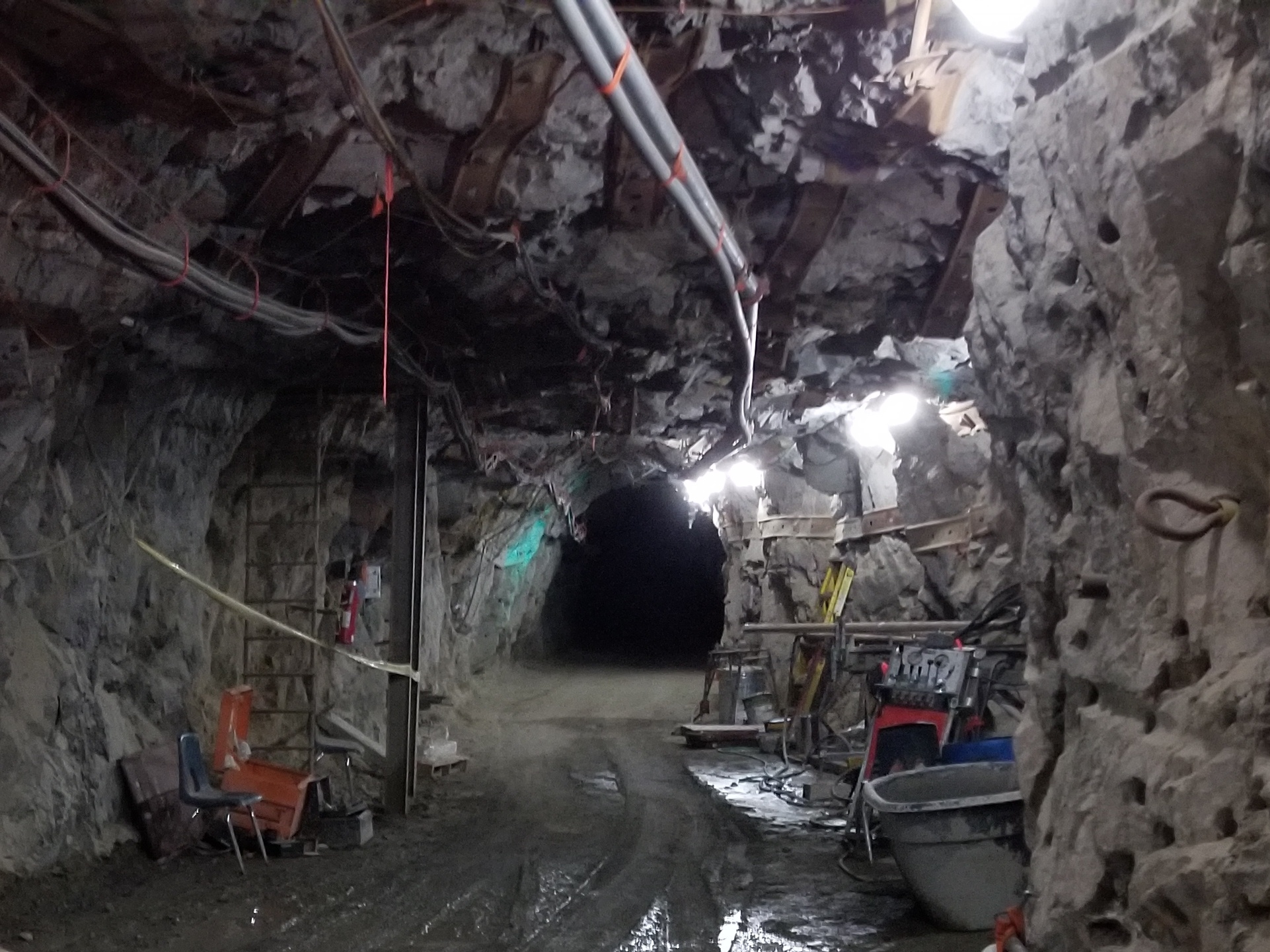} &
\includegraphics[width=0.3\textwidth]{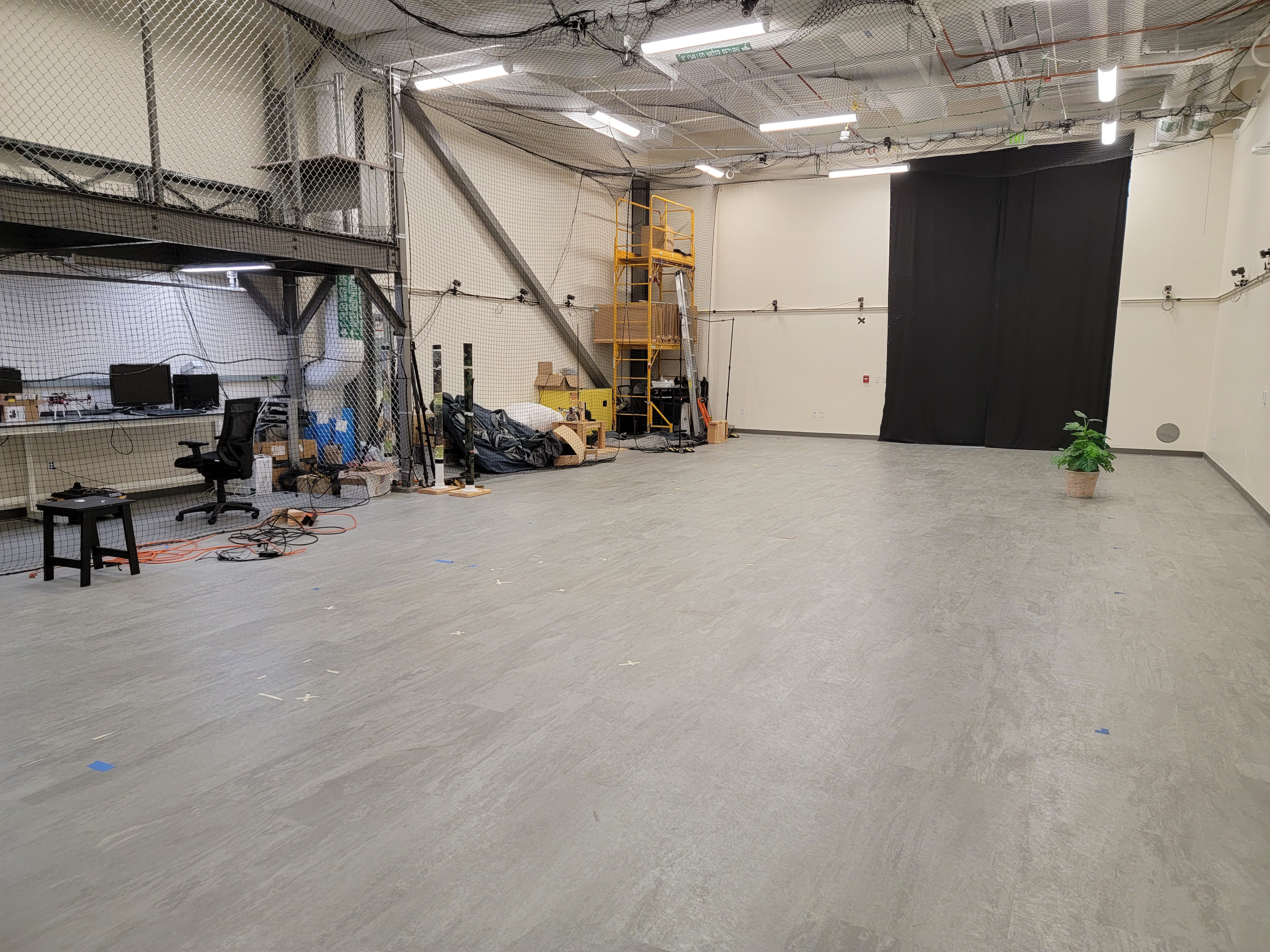} &
\includegraphics[width=0.3\textwidth]{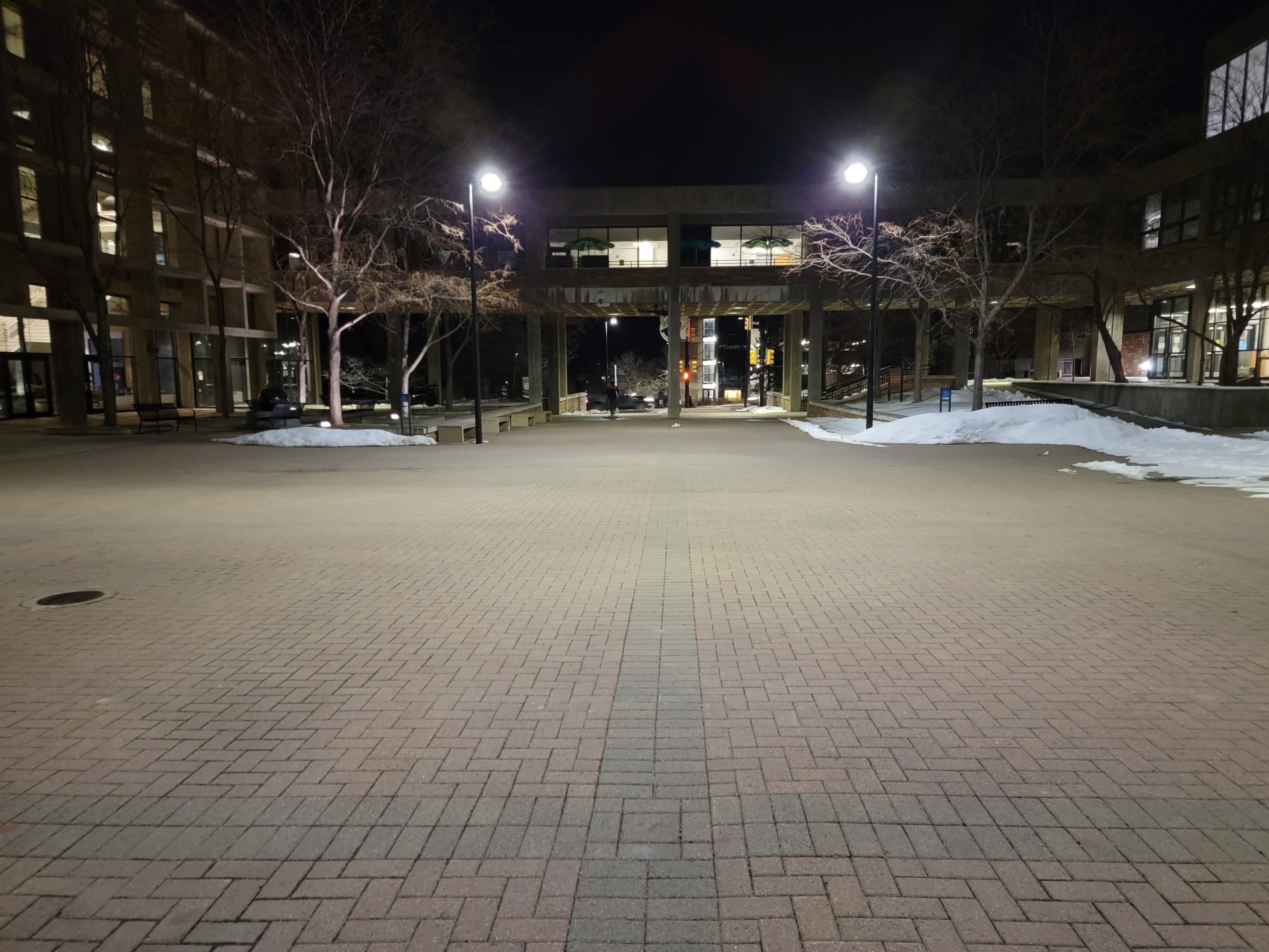}\\
\textbf{(d)}  & \textbf{(e)}  & \textbf{f} \\[6pt]
\end{tabular}
\caption{ Examples of various environments this dataset covers including \textbf{(a)} Intelligent Robotics Lab
\textbf{(b)} Engineering Center hallways
\textbf{(c)} Boulder Creek Path where the longboard data was captured
\textbf{(d)} Edgar Mine
\textbf{(e)} Autonomous Systems Programming Evaluating and Networking Lab motion capture space
and \textbf{(f)} Engineering Center courtyard where the outdoor data was captured. }
\label{fig:locations}
\end{figure*}

Data was collected in several unique environments with the goal of providing a diverse range of sensor data. The Intelligent Robotics Laboratory (IRL) and engineering center (EC) hallways sequences were taken in an indoor built environment and the EC courtyard sequences capture data in an outdoor built environment. In the Boulder Creek Path sequences the sensor rig moves rapidly through an outdoor environment with dynamic obstacles including pedestrians, cyclists, and road vehicles. There are two sets of sequences taken in Edgar Experimental Mine, a former silver and gold mine managed by Colorado School of Mines: Edgar Army sequences were taken in large, open areas of the mine and the Edgar Classroom sequences were taken in the mine's tighter, more constrained passageways. The ASPEN Lab sequences were taken in a motion capture space with a variety of obstacles.

Example images of the various environments are shown in Figure \ref{fig:locations}. The amount of data collected at each location, along with their associated location file names are shown in Table \ref{tab:timings}. 

\begin{table}
\centering
 \begin{tabular}{| p{3.0cm} p{2.0cm} p{2.0cm}  |} 
 \hline
 Location & File Root & Time (mm:ss) \\ [0.5ex]
\hline\hline 
 IRL & arpg\_lab & 10:24 \\ [0.5ex]
 \hline
 EC Hallways & ec\_hallways & 15:07 \\ [0.5ex]
 \hline
 Boulder Creek Path & longboard & 31:53 \\ [0.5ex]
 \hline
 Edgar Army & edgar\_army & 25:20 \\  [0.5ex] 
 \hline
 Edgar Classroom & edgar\_classroom & 19:20 \\ [0.5ex]
 \hline
 ASPEN Lab & aspen & 21:14 \\ [0.5ex]
 \hline
 EC Courtyard & outdoors & 19:32 \\ [0.5ex]
 \hline
\end{tabular}
\caption{Amount of data in time collected at each location} \label{tab:timings}
\end{table}

\begin{figure}
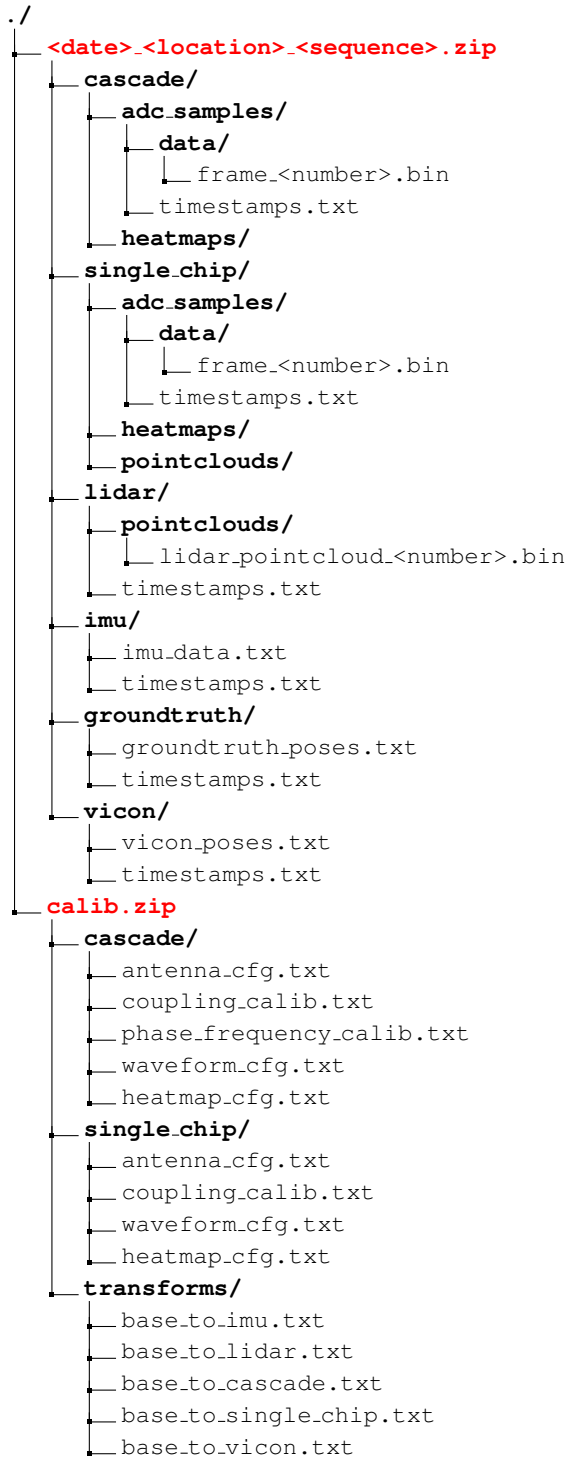

\small
\dirtree{%
.1 \textbf{./}.
.2 \textcolor{red}{\textbf{<date>\_<location>\_<sequence>.zip}}.
.3 \textbf{cascade/}.
.4 \textbf{adc\_samples/}.
.5 \textbf{data/}.
.6 frame\_<number>.bin.
.5 timestamps.txt.
.4 \textbf{heatmaps/}.
.3 \textbf{single\_chip/}.
.4 \textbf{adc\_samples/}.
.5 \textbf{data/}.
.6 frame\_<number>.bin.
.5 timestamps.txt.
.4 \textbf{heatmaps/}.
.4 \textbf{pointclouds/}.
.3 \textbf{lidar/}.
.4 \textbf{pointclouds/}.
.5 lidar\_pointcloud\_<number>.bin.
.4 timestamps.txt.
.3 \textbf{imu/}.
.4 imu\_data.txt.
.4 timestamps.txt.
.3 \textbf{groundtruth/}.
.4 groundtruth\_poses.txt.
.4 timestamps.txt.
.3 \textbf{vicon/}.
.4 vicon\_poses.txt.
.4 timestamps.txt.
.2 \textcolor{red}{\textbf{calib.zip}}.
.3 \textbf{cascade/}.
.4 antenna\_cfg.txt.
.4 coupling\_calib.txt.
.4 phase\_frequency\_calib.txt.
.4 waveform\_cfg.txt.
.4 heatmap\_cfg.txt.
.3 \textbf{single\_chip/}.
.4 antenna\_cfg.txt.
.4 coupling\_calib.txt.
.4 waveform\_cfg.txt.
.4 heatmap\_cfg.txt.
.3 \textbf{transforms/}.
.4 base\_to\_imu.txt.
.4 base\_to\_lidar.txt.
.4 base\_to\_cascade.txt.
.4 base\_to\_single\_chip.txt.
.4 base\_to\_vicon.txt.
}
\caption{Structure of the zip files for each sequence in the dataset.}
\label{fig:directory_chart}
\end{figure}

All sensor readings for a given sequence are grouped in a single compressed file named \texttt{<date>\_<location>\_<sequence>.zip}, where \texttt{<date>}, \texttt{<location>}, and \texttt{<sequence>} are placeholders for the recording date, location, and sequence number respectively. The directory structure is shown in Figure \ref{fig:directory_chart}.

Timestamps for each sensor reading are stored in a \texttt{timestamps.txt} file for each sensor. Each line in \texttt{timestamps.txt} gives the time of a sensor measurement in seconds. 

As previously stated we provide three levels of processing for radar data: unprocessed ADC samples, 3D range-azimuth-elevation intensity-Doppler velocity heatmaps, and target detections from the single-chip radar sensor. The cascaded sensor does not do any signal processing or target detection onboard, so point targets are not provided for the cascade sensor. By providing these different data types will enable users to develop application specific methods for signal processing, target detection, scan alignment, etc. 

\subsubsection{Binary File Format}

Radar and lidar sensor measurements are very large. Storing these measurements in human-readable format would be impractical due both to their size and the time required to access the data. So for these sensors we use a simple binary encoding that can be read in any language without external dependencies. Each binary file stores an array of numeric values in little-endian order. To obtain the array one only needs to read the file's contents into a byte array and then convert to the appropriate type with \texttt{memcpy} or similar. The organization of the values in the resulting array differs for each measurement type and is described in detail below. We also provide both Python and MATLAB scripts demonstrating correct parsing of the binary files for each sensor type.


\subsubsection{Raw Radar Data}

Each frame of raw radar data is stored in a binary file \texttt{frame\_<number>.bin}. Each of these files stores an array of 16 bit signed integer values. The values that make up the ADC sample for sample index $s$, chirp index $c$, receiver index $r$, and transmitter index $t$ are indexed as 

\begin{equation}
    i = 2(s + N_s(c + N_c(r + tN_r)))
\end{equation}

\noindent the value at index $i$ make is the real (or I) component and value at index $i+1$ is the imaginary (or Q) component. Python and MATLAB examples are provided to demonstrate correct parsing of these binary files.

\subsubsection{Radar Heatmaps}

Each radar heatmap is stored in a binary file \texttt{heatmap\_<number>.bin}. Each of these files stores an array of 32 bit floating point values. The heatmaps have dimension $N_e\times N_a\times N_r \times 2$ where $N_e$ is the number of elevation bins in the heatmap, $N_a$ is the number of azimuth bins, and $N_r$ is the number of range bins. The Doppler velocity dimension in the heatmaps has been collapsed, and the return in each range, azimuth, and elevation bin is represented by 2 32 bit floating point values: the intensity of the peak in the Doppler velocity spectrum for that location, and the range rate of that peak in meters per second. Negative range rates indicate motion toward the sensor and positive range rates indicate motion away from the sensor. The values that make up the intensity and range rate for range index $r$, azimuth index $a$, and elevation index $e$ are indexed as

\begin{equation}
    i = 2(r + N_r(a + eN_a))
\end{equation}

\noindent where the value at index $i$ is the intensity value and the value at index $i+1$ is the range rate. As with the raw radar data, Python and MATLAB examples are provided to demonstrate correct parsing.

\subsubsection{Radar Pointclouds}

Each radar pointcloud is stored in a binary file \texttt{radar\_pointcloud\_<number>.bin}. Each of these files stores an array of 32 bit floating point values. Each radar point consists of 5 contiguous values: its $(x,y,z)$ location in meters in the sensor's local frame, its intensity value, and its range rate in meters per second. As with the other data types, Python and MATLAB examples are provided to demonstrate correct parsing. The number of points per scan is not constant and depends on the number of detected targets in the scan, but there are generally less than 300 points per scan.

\subsubsection{Lidar Pointclouds}

Each lidar pointcloud is stored in a binary file \texttt{lidar\_pointcloud\_<number>.bin}. As with radar pointcloud files, each lidar pointcloud file stores an array of 32 bit floating point values. Each lidar point consists of 4 contiguous values: its $(x,y,z)$ location in meters in the sensor's local frame and its intensity. There are 65,536 points per lidar scan. Python and MATLAB examples are provided to demonstrate correct parsing.

\subsubsection{IMU}

All IMU measurements are stored in a single text file, \texttt{imu\_data.txt}, with one measurement per line. Each measurement consists of 6 space-delimited floating point values: the linear accelerations $(a_x,a_y,a_z)$ followed by the angular rates $(\alpha_x,\alpha_y,\alpha_z)$. Python and MATLAB examples are provided to demonstrate correct parsing.

\subsubsection{Groundtruth}

Groundtruth poses of the sensor rig are generated from a globally optimized pose graph with IMU, lidar, and loop closures as constraints using the method described in \cite{cartographer}. All groundtruth poses for a sequence are given in a single text file \texttt{groundtruth\_poses.txt} with one pose per line. Each pose consists of 7 space-delimited floating point values: the position $(x,y,z)$ followed by an orientation quaternion $(x,y,z,w)$. Groundtruth poses are given relative to the starting pose of the sequence. Python and MATLAB examples are provided to demonstrate correct parsing.

\subsubsection{Vicon}

All vicon poses for a sequence are given in a single text file \texttt{vicon\_poses.txt} with one pose per line. As with the lidar-based groundtruth files, each pose consists of 7 space-delimited floating point values: the position $(x,y,z)$ followed by an orientation quaternion $(x,y,z,w)$. Vicon poses are given relative to the vicon system's coordinate frame. Vicon poses are only available for sequences in the ASPEN lab. Python and MATLAB examples are provided to demonstrate correct parsing.

\subsection{Development Tools}

We provide development tools for manipulating our raw data on the ColoRadar website. These include example Python tools for reading and plotting the provided data files. These are described in detail in the \texttt{README.md} file provided with the development tools. Here we will briefly describe the tools and their functions. \texttt{dataset\_loaders.py} provides functions to load sensor measurements and calibrations from dataset files. These functions parse the raw data and organize it into logical data structures. For instance, the function \texttt{get\_heatmap()} reads bytes from a radar heatmap file, converts those bytes to intensity and range rate values, and returns them in a numpy array of dimension $N_e \times N_a \times N_r \times 2$. The function \texttt{get\_cascade\_params()} reads config files for the cascaded radar sensor and returns a dictionary where each key-value pair is a parameter for the radar sensor. \texttt{plot\_pointclouds.py} demonstrates how the functions in \texttt{dataset\_loaders.py} can be used to create an animation of the sensor readings and groundtruth poses in a sensor-centric reference frame. MATLAB equivalents of these tools are also provided.

\section{Sensor Calibration}

\subsection{Radar Calibration}

Four types of calibration parameters are provided for the radar sensor data. The first is a mapping of the location of the TX and RX antennas on each radar board. These are stored in text files found in \texttt{calib/cascade/antenna\_cfg.txt} and \texttt{calib/single\_chip/antenna\_cfg.txt}.

The second type of calibration is the waveform parameters. These include the start frequency, frequency slope, ADC sample rate, and other parameters necessary for signal processing. These parameters are found in \texttt{calib/cascade/waveform\_cfg.txt} and \texttt{calib/single\_chip/waveform\_cfg.txt}.

The third type is the antenna coupling calibration. Antenna coupling occurs when a current in one antenna element unintentionally induces a voltage in adjacent elements. Antenna coupling calibration compensates for this as

\begin{equation}
    \mathbf{R}_{rca}^{\text{calibrated}} = \mathbf{R}_{rca}^\text{raw} - \mathbf{C}_{rca}
\end{equation}

\noindent where $\mathbf{R}_{rca}^\text{raw}$ is the uncalibrated range FFT output for range index $r$, chirp index $c$, and virtual antenna index $a$, $\mathbf{R}_{rca}^\text{calibrated}$ is the corresponding calibration result, and $\mathbf{C}_{rca}$ calibration value, i.e. the range FFT result at $r$, $c$, and $a$ when no targets are present.

Antenna coupling calibration files are provided in \texttt{calib/cascade/coupling\_calib.txt} and \texttt{calib/single\_chip/coupling\_calib.txt} and Python code is provided to show how to apply the calibration.

The fourth type is phase and frequency calibration. These parameters are needed for the cascaded sensor only. This is because the cascaded sensor uses 4 separate chips to control different parts of the antenna array and there may be small mismatches in the frequency, phase, and amplitude across these 4 chips. Phase and frequency calibration corrects for this.

In frequency calibration, we start with a calibration vector $\mathbf{f}$ of dimension $N_\text{rx}N_\text{tx}$. From this we calculate a frequency calibration matrix $\mathbf{F}$ of dimension $N_\text{rx}N_\text{tx}\times N_\text{range}$ as

\begin{equation}
    \mathbf{F} = 2\pi \mathbf{f} \frac{f_\text{s}^\text{cal}f_\text{slope}}{f_\text{s}f_\text{slope}^\text{cal}}\times \mathbf{n}
\end{equation}

\noindent where $f_\text{s}$ is the ADC sampling rate, $f_\text{s}^\text{cal}$ is the sampling rate used in calibration, $f_\text{slope}$ is the frequency slope, $f_\text{slope}^\text{cal}$is the frequency slope used in calibration, and $\mathbf{n}=[0,N_\text{range})$. The raw ADC samples for each chirp $\mathbf{A}_c$ are then calibrated as

\begin{equation}
    \mathbf{A}_c^\text{fcal} = \mathbf{A}_c \odot e^{-j\mathbf{F}}
\end{equation}

\noindent where $\odot$ denotes elementwise multiplication. Phase calibration is somewhat simpler. In phase calibration we start with a matrix $\mathbf{P}$ of dimension $N_\text{rx}\times N_\text{tx}$. $\mathbf{P}$ is applied to $\mathbf{A}_{rc}^\text{fcal}$, the frequency calibrated ADC samples from each antenna pair at range index $r$ and chirp index $c$ as 

\begin{equation}
    \mathbf{A}_{rc}^\text{pcal} = \mathbf{A}_{rc}^\text{fcal} \oslash \mathbf{P}
\end{equation}

\noindent where $\oslash$ denotes elementwise division.

Phase and frequency calibration vectors can be found in \texttt{calib/cascade/phase\_frequency\_calib.txt} and Python code is provided to show how to apply the calibration. 

\subsection{Extrinsic Calibration}

Extrinsic transforms between sensors are given relative to the sensor rig's base frame. The sensor rig's base frame is coincident with the IMU in translation and oriented to align with the ENU frame ($x$ right, $y$ forward, $z$ up). Transformations between the base frame and each sensor are given in \texttt{calib/transforms/base\_to\_<sensor>.txt}. Each of these files has two lines. The first gives the translation $(x,y,z)$ and the second gives the rotation as a quaternion in the order $(x,y,z,w)$. The extrinsic transformations for all of the sensors are hand measured. The transformation between the lidar and radar sensors is further refined via the following process: we took a dataset in which the sensor rig was held static and two obstacles are clearly visible in the lidar pointcloud and the heatmaps for the single chip and and cascaded radar sensors. The extrinsic transformations for the lidar and single chip sensors are then manually adjusted until the locations of the obstacles in the lidar pointcloud and single chip heatmap coincide with those in the cascaded heatmap. The extrinsic transforms do not change appreciably between runs, so we include only one set of extrinsic transforms. 




\section{Summary and Future Work}

In this paper we have presented a dataset focused on robotic perception using millimeter wave radar. We provide data in a diverse range of environments including indoor, outdoor, and subterranean environment and we include motion in 6DoF. We also include 4D radar measurements from two different sensors in three different formats. These factors make this dataset highly useful in research for autonomous robotics tasks such as odometry, mapping, and SLAM in diverse roles from disaster response to subterranean exploration. Currently, there is great interest in radar-based 3D odometry, mapping, and SLAM in the research community but very few published methods exist. The availability of a diverse dataset focused on these tasks will help spur development in this area. The usefulness of the dataset could be extended to object classification and semantic segmentation by adding semantic labeling to the lidar pointclouds. This is another potential direction for future work.

\bibliographystyle{SageH}
\bibliography{references}

\end{document}